\documentclass[sigconf]{acmart}

\usepackage[T1]{fontenc}  
\usepackage{hyperref} 
\usepackage{url}  
\usepackage{booktabs} 
\usepackage{amsfonts} 
\usepackage{nicefrac} 
\usepackage{microtype}  
\usepackage{xcolor} 
\usepackage{multirow}
\usepackage{anyfontsize}
\usepackage{bm} 
\usepackage{array}
\usepackage{enumitem}
\usepackage{graphicx}
\usepackage{tablefootnote}
\usepackage{latexsym}
\usepackage{amsmath}

\usepackage{amssymb}
\usepackage{mathtools}
\usepackage{amsthm}
\theoremstyle{plain}
\newtheorem{theorem}{Theorem}[section]
\newtheorem{proposition}[theorem]{Proposition}

\theoremstyle{definition}
\newtheorem{definition}[theorem]{Definition}
\newtheorem{assumption}[theorem]{Assumption}
\theoremstyle{remark}

\usepackage[ruled]{algorithm2e}
\usepackage{balance}

\AtBeginDocument{%
  }

\setcopyright{acmcopyright}
\copyrightyear{2025}
\acmYear{2025}
\setcopyright{acmlicensed}\acmConference[WSDM '25]{Proceedings of the Eighteenth ACM International Conference on Web Search and Data Mining}{March 10--14, 2025}{Hannover, Germany}
\acmBooktitle{Proceedings of the Eighteenth ACM International Conference on Web Search and Data Mining (WSDM '25), March 10--14, 2025, Hannover, Germany}
\acmDOI{10.1145/3701551.3703568}
\acmISBN{979-8-4007-1329-3/25/03}

\begin{document}
\title{Heterophilic Graph Neural Networks Optimization with Causal Message-passing}

\author{Botao Wang}
\email{bwangbk@connect.ust.hk}
\additionalaffiliation{Hong Kong University of Science and Technology (Guangzhou)}
\affiliation{
  \institution{Hong Kong University of Science and Technology}
  \city{Hong Kong SAR}
  \country{China}
}

\author{Jia Li}
\authornote{Corresponding author.}
\email{jialee@ust.hk}
\affiliation{
  \institution{Hong Kong University of Science and Technology (Guangzhou)}
  \city{Guangzhou}
  \country{China}
}

\author{Heng Chang}
\email{changh.heng@gmail.com}
\affiliation{
  \institution{Huawei Technologies Co., Ltd}
  \city{Beijing}
  \country{China}
}

\author{Keli Zhang}
\email{zhangkeli1@huawei.com}
\affiliation{
  \institution{Huawei Noah's Ark Lab}
  \city{Shenzhen}
  \country{China}
}

\author{Fugee Tsung}
\authornotemark[1]
\email{season@ust.hk}
\affiliation{
  \institution{Hong Kong University of Science and Technology}
  \city{Hong Kong SAR}
  \country{China}
}

\renewcommand{\shortauthors}{Botao Wang, Jia Li, Heng Chang, Keli Zhang and Fugee Tsung}

\begin{abstract}
  In this work, we discover that causal inference provides a promising approach to capture heterophilic message-passing in Graph Neural Network (GNN). By leveraging cause-effect analysis, we can discern heterophilic edges based on asymmetric node dependency. The learned causal structure offers more accurate relationships among nodes. To reduce the computational complexity, we introduce intervention-based causal inference in graph learning. We first simplify causal analysis on graphs by formulating it as a structural learning model and define the optimization problem within the Bayesian scheme. We then present an analysis of decomposing the optimization target into a consistency penalty and a structure modification based on cause-effect relations. We then estimate this target by conditional entropy and present insights into how conditional entropy quantifies the heterophily. Accordingly, we propose CausalMP, a causal message-passing discovery network for heterophilic graph learning, that iteratively learns the explicit causal structure of input graphs. We conduct extensive experiments in both heterophilic and homophilic graph settings. The result demonstrates that the our model achieves superior link prediction performance. Training on causal structure can also enhance node representation in classification task across different base models.
\end{abstract}


\maketitle

\section{Introduction}
The message-passing mechanism of graph neural network (GNN) inherently assumes homophily, which degrades in the heterophilic graphs. Heterophily refers to the characteristic of graphs whose edges are more likely to connect the nodes from different classes. In such graphs, the representations of node pairs become less distinguishable after being smoothed by their neighborhood features. This issue becomes particularly pronounced when there is lack in the node label information, such as link prediction task and few-shot node classification. The heterophilic edges can also act as the noise that hinder the optimization. Numerous studies have been proposed to improve message passing, aiming to learn more fair representations that are not affected by the heterophilic edges. They are usually dedicated to disentangling heterophilic information \cite{abu2019mixhop,chien2020adaptive,yan2022two} or improving the information gathering process in the graph \cite{kong2023goat,zhou2022greto,wang2022acmp,zheng2023finding}. However, these techniques are task-specific with low generalization ability. We also find most fail to demonstrate substantial performance improvements across both homophilic and heterophilic graphs.

\begin{figure}[t]
  \centering
  \includegraphics[width=0.8\columnwidth]{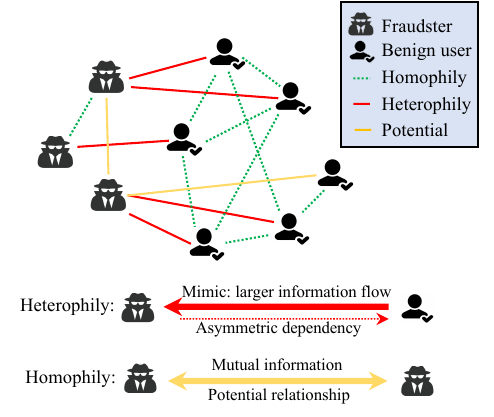}
  \caption{Detect heterophily by causal-effect estimation from asymmetric information flow that results from the mimic behaviors of fraudsters. }
  \Description{Detect heterophily by causal-effect estimation from asymmetric information flow that results from the mimic behaviors of fraudsters}
  \label{fig:intro}
\end{figure}

Causal inference emerges as a promising approach for capturing the cause-effect among variables according to the distribution of observed data. While self-training is a widely used and straightforward strategy for relationship discovery \cite{wang2023deep,cheng2023wiener}, it focuses on strengthening correlations, which is inadequate for complex and heterophilic graphs. Causal inference can detect dependencies at a higher level \cite{Cotta2023Causal,zhao2022learning}. It can be utilized to identify heterophilic edges, which exhibit asymmetric dependencies between connected node pairs.
To illustrate the concept, we consider a camouflaged fraudster detection case, as depicted in Fig.\ref{fig:intro}. The nodes consist of fraudsters and benign users. Fraudsters can mimic the behaviors of benign users and establish connections with them, having similar node and structural embeddings. These heterophilic links make it challenging for GNN-based models to effectively detect fraudsters. However, there exists an information flow from benign users to fraudsters due to the mimicry behavior. Causal inference can detect the node dependencies where the fraudsters depend on the connected benign users, which suggest the presence of heterophily. If we revise the distinguished heterophilic edges between directed edges from benign users to fraudsters, the graph can mitigate the feature smoothing. The updated graph aligns more closely with the true information flow and improves the aggregation process within GNNs.

Causal inference typically involves conducting independence tests to determine the sub-structure of the causal graph. According to Pearl's causal analysis scheme \cite{pearl2009causality}, there are two higher levels than association (correlation) analysis: intervention-based and counterfactual inference. Although counterfactual inference is at the highest level, it is request non-existed knowledge that usually estimated by a generative process or searching for an estimation unobtainable data \cite{yao2021survey}, which is time-consuming and computationally expensive. The intervention-based method can be realized by graph augmentation to uncover variable dependency \cite{vu2020pgm}. It does not require background knowledge and can improve the information aggregation process \cite{luan2022revisiting}. The nature of the variable distribution is described by the node dependencies, whose inference process should ideally have better generalization ability. And the learned causal structure should be effective across heterophilic and homophilic graph.

However, there remains challenges of application of causal inference in GNNs. First, most causal inference methods for graph data primarily focus on graph classification \cite{wu2022discovering, fan2022debiasing, bevilacqua2021size}or invariant learning \cite{lin2021generative,lin2022orphicx}. The analysis of causality among node variables for a single input graph remains undefined. Besides, it is crucial to construct a message passing that maintains generalization ability. The learned causal structure should align with node relationships, enhancing the performance of GNNs across various downstream tasks and graphs. Furthermore, to the best of our knowledge, there have been no studies that apply causal inference specifically to address heterophily. To tackle the challenges above, we propose a network that learns explicit \underline{Causal} \underline{M}essage-\underline{P}assing (CausalMP) of input graph by node dependency discovery. We quantify the intervention-based cause-effect by the dependencies between node pairs. Specifically, we construct a causality estimator among the node variables by comparing the conditional entropy in both directions of the edge. We demonstrate its effectiveness in indicating the presence of heterophily. Then, we modify the topology accordingly for the subsequent optimization. Experiments show that the proposed model achieves better performance in link prediction across both heterophilic and homophilic graphs. Moreover, the learned causal structure can improve the GNNs' training in node classification performance of different baseline methods, even when there are limited labeled samples.

\section{Preliminaries}
\textbf{$do$-operator for causal inference.} In causal inference, the $do$-operator is usually applied to evaluate the cause-effect relation. It represents interventions conducted on variables to assess their causal effects. By applying interventions, we can quantify the impact of intervened variables by observing the resulting variations in the distribution of other variables. We use $P(X_2|\text{do}(X_1))$ to describe the probability density of $X_1\rightarrow X_2$, which quantifies the effect on $X_2$ resulting from $X_1$. $do$-operator can be calculated by conditional probability in some scenarios. If there are three variables $X_1,X_2,X_3$ and we wish to use the individual treatment effect (ITE) to quantify the effect of $X_1$ on $X_2$, we can apply the $do$-operator as:
\begin{equation}
  P(X_2|\text{do}(X_1))=\sum_{X_3} P(X_2|X_1,X_3)P(X_3).
\end{equation}

If $X_1$ is binary treatment, then the ITE is calculated by:
\begin{equation}
  ITE(X_1\rightarrow X_2)=P(X_2|\text{do}(X_1=1))-P(X_2|\text{do}(X_1=0)).
\end{equation}

In the multivariable situations, $X_i, i=1,2,...$ can represent variable groups. Then the structural equation model (SEM) can be employed for causal analysis. In SEM, the relationships among variables are depicted using directed edges in a graphical model denoted as $\mathcal{M}(X,A_c)$, where $X$ represents the variables and $A_c$ represents the causal graph in the form of an adjacency matrix. The model $\mathcal{M}$ typically starts from either a fully connected graph or an initialization based on prior knowledge graph, whose adjacency matrix denoted by $A$. Various algorithms, such as PC, FCI, \emph{etc.}, have been proposed to search for subgraphs that are faithful to the true causal graph $A_c$. Subsequently, conditional independence tests are conducted on the observed data to maximize the averaged treatment effect (ATE) after pruning.
\begin{equation}\label{eq:ate}
  ATE=\mathbb{E}_{(i,j)\in A} [ITE(X_i\rightarrow X_j)].
\end{equation} 
\textbf{Estimation of causality.} Intervention-based methods in structural equation modeling (SEM) exhibit superior performance in independence tests for uncovering causal relationships. However, these methods also demand greater computational resources. One such intervention-based analysis involves estimating the joint causal distribution of variables \cite{pearl2010causal} and maximizing their likelihood \cite{van2010targeted}.
If the joint distribution of observed data can be factorized explicitly, the intervened SEM $\mathcal{M}_{\text{do}(I)}$ can be written as:
\begin{equation}\label{eq:int_dist}
  P_{\text{do}(I)}(X)=\prod\limits_{i\in I} \left. P(X_i|Pa_i^{A_c})\right\vert_{X_I=x_I}
\end{equation}
where $Pa_i^{A_c}$ is the parent nodes of node $i$ in graph $A_c$, $X_I$ is the intervened variable.

In the Bayesian scheme, the distribution of the potential causal graphs can be denoted by $P(A_c), A_c\in\mathcal{A}$. With the observed graph data, we can represent the prior of the causal graph's parameters as $P(\theta_c|A_c), \theta_c\in\Theta_c$. Then the prior belief is written as $P(A_c,\theta_c)=P(\theta_c|A_c)P(A_c)$. We define the likelihood of the node features as $P(X|\theta_c,A_c)$. The marginal likelihood is calculated by:
\begin{equation}\label{eq:bayes}
  P(X|A_c) = \int_{\Theta_c} P(X|\theta_c,A_c)P(\theta_c|A_c) d\theta_c.
\end{equation}
Then, it can be estimated by the Monte Carlo algorithm according to the specific model via the assumptions in the specific tasks.

\textbf{Heterophilic graph.} Generally, heterophilic graph refers to the edge heterophily \cite{dai2022label}, where the edges often link nodes with different labels. We can use homophily ratio \cite{zhu2020beyond} as the metric. Given a graph $G(\mathcal{V},\mathcal{E})$, its homophily ratio is 
$R_h = |\{(u,v)\in\mathcal{E}:y_u=y_v\}|/|\mathcal{E}|$,
where $\mathcal{V}$ is the node set, $\mathcal{E}$ is the edge set, $y_u,y_v$ are the labels of node $u,v$ respectively. $R_h$ ranges from 0 to 1. A value close to 1 implies strong homophily, while a value close to 0 indicates strong heterophily.

\section{Causal Inference on Heterophily}
The edges in the graph serve as indicators of association among the nodes. They can be treated as prior knowledge or skeletons for causal analysis. By optimizing the graph structure to align with the causal structure, we detect the heterophily and enhance the information-gathering of GNNs. This, in turn, aids GNNs in developing a better understanding of the node relationships during link prediction.

\subsection{Information aggregation in GNN}
In this Section, we start by forming the causal inference of the node dependency problem in GNNs. We present the assumptions regarding the properties of causality in the context of GNNs. 
Given a graph $G(X,A)$ with $N$ nodes, the features on each node $X_i\in\mathbb{R}^D, i\in [N]$ are considered as a group of variables. The adjacency matrix $A\in\mathbb{R}^{N\times N}$ is the initialization of causal structure. A graph convolution layer can be denoted by:
\begin{equation}
  \hat{X}=\text{AGG}(A,X)=\sigma(WAX+b)\triangleq \sigma\left(\Gamma X+b\right),
\end{equation}
where $\text{AGG}(\cdot)$ is the aggregation function, $W,b$ are the weight matrix and bias vector of the GNN layer, $\sigma(\cdot)$ is the activation function, and $\Gamma$ is the connection coefficient matrix of the layer. $\Gamma$ refers to a quantification of the cause-effect, where $\Gamma_{jk}=0$ if $X_k\notin Pa^{A_c}(X_j)$, where $Pa^{A_c}(\cdot)$ is the parents set of the given node in adjacency matrix $A_c$.

Local Markov property is a commonly applied assumption in the causal structure learning \cite{pellet2008using}. It enables  the implied conditional independencies being read off from a given causal structure \cite{kalisch2014causal}. While, enumerate all the conditions in GNN is NP-hard problem. Thus, we focus on the most primary cause-effect relationships in each convolution layer. For a node variable $X_i$, we only consider the 1-hot neighbors and ignore the spouse nodes in Markov blanket, which is a common simplification in GNN. Specifically,  we assume it is conditional independent of the rest of its neighbors. Then $X_i\perp X_{\setminus i\setminus X(i)}|X_{\mathcal{N}(i)}$, where $\mathcal{N}(i)$ are the neighbor nodes of node $X_i$ in the input graph.

In local perspective of GNN, a center node $X_i$, denoted as $Y$ following, acquires information from parent nodes $X_j \in Pa^{A}(X)$. Causal structure estimation is to identify its \textit{cause}, represented by $\Gamma_{ij} \neq 0$. Although ground truth of causal structure is often unfeasible in practice, we can still learn invariant causal connections of center node, which possesses optimal generalization capabilities. Subsequently, we can make following assumption to provide an explicit definition of causality for node variables of the input graph.

\begin{assumption}\textbf{Aggregation Invariant.} 
For the current observed causal connections $\Gamma=[\Gamma_1,...,\Gamma_N]\in\mathbb{R}^{D\times N}$, there exists a optimal subset $S^*={k:\Gamma_k\neq\textbf{0}}\subseteq\{1,...,N\}$, that satisfies in any context $\xi\in\Xi$
\begin{equation}\label{eq:assumption}
  \begin{aligned}
  X^\xi &=h_x(C_f^\xi,\eta)+E_x^\xi \\
  Y^\xi &=h_y(\Gamma X^\xi,C_f^\xi,\epsilon)+E_y^\xi,
  \end{aligned}
\end{equation}
where $X,Y$ are the node feature variables, $C_f$ is the confounder, $h_x(\cdot),h_y(\cdot)$ are the effect functions of the confounder on $X,Y$, and $\eta,\epsilon$ are the random noise with mean of 0, $E_x,E_y$ are the random bias vectors. Here, $C_f,E_x,E_y$ are jointly independent.
\label{assump:aggr_inv}
\end{assumption}

The aforementioned assumption implies the conditional distribution $Y^\xi|X_{S^*}^\xi,C_f^\xi$ for any given context, where each context corresponds to an intervention-based independence experiment. By making this assumption, we consider the existence of common knowledge that holds true across contexts, representing the causal connections. Besides, when applying an embedding model, the assumption guarantees the stability of the learned embeddings, ensuring a consistent joint entropy after graph augmentation. It is important to note that the causal connection $S^*$ is not necessarily unique, indicating that the learned causal structure of the original graph can be variant.

\subsection{Estimation of Intervention-based causality}

In the Bayesian framework, the optimization involves maximizing the margin likelihood as Eq.(\ref{eq:bayes}). We aim to learn the parameter of the causal structure $\theta$. The posterior of the final causal structure $A_c$ and its parameters are denoted by $\theta_c$ are $P(A_c|X) \propto P(A_c)P(X|A_c) $ and $P(\theta_c|X,A_c)\propto P(\theta_c|A_c)P(X|\theta_c,A_c)$. They can be estimated by $P(A|X), P(\theta_c|X,A)$ respectively after initialization. To approximate the optimization, we conduct Bayesian experimental design (BED) \cite{Kugelgen2019Optimal} to model the node dependency in the form of entropy. Then utilize Monte Carlo estimator \cite{Kugelgen2019Optimal} for the $do$-intervention to quantify the point-wise causal relationship. The optimization problem is formalized as the following proposition.

\begin{proposition}
  Given the intervention strategy $I\in\mathcal{I}$, if we have the condition distribution $P(X_{-I},A|X), P(A|X)$, the causal structure $A_c$ and corresponding optimal intervention target $x_I$ can be obtained by optimizing:
  \begin{equation}\label{eq:estimated}
  x_I,A_c = \arg\max_{x_I,A}\mathbb{E}_{P(X_{-I}|do(X_I=x_I))}[
  \mathbf{H}(P(X_{-I},A|X))-\mathbf{H}(P(A|X))]
  \end{equation}
\label{prop:estimated}
where $X_I$ are the intervened node features, $X_{-I}$ are the non-intervened features, $\mathbf{H}(\cdot)$ is the entropy.
\end{proposition}

To approximate the optimization, the Bayesian experimental design (BED) approach can be employed, as described in \cite{Kugelgen2019Optimal}. BED utilizes the uncertainties associated with the causal structure, which are captured by the posterior distribution. As such, we have the following proposition:

\begin{proposition} The optimization problem for the causal inference of node dependency in a graph $G(X,A)$ on GNN-based embedding networks $f:G\rightarrow B$ can be formulated as:
\begin{equation}\label{eq:optim}
  \begin{aligned}
  \xi^*, \theta_c =& \arg \max_{\xi,\theta} \int U(B|\xi)P(B|\xi)dB, \\
 U(B|\xi) =&\int_\Theta P(\theta|B,\xi)\log P(\theta|B,\xi)d\theta-\int_\Theta P(\theta) \log P(\theta)d\theta,
  \end{aligned}
\end{equation}
where $\xi^*$ is the optimal intervention experiment, $\theta_c\in \Theta$ is the parameter of the causal structure with a prior $P(\theta)$, $P(B|\xi)$ is the posterior of embeddings given the experiment $\xi$, and $U(B|\xi)$ is the utility function that quantifies the information gain of the causal structure.
\label{prop:optim}
\end{proposition}

According to the proposition, the learning of the causal structure can be achieved by maximizing the usefulness metric $U(B|\xi)$. And the optimization process simultaneously improves the posterior probability $P(B|\xi)$.

In the graph data, we transfer the optimization problem into Eq.(\ref{eq:optim}). To proceed with this optimization, we require an estimation of the utility function based on the definition of causality.
In the point-wise relationship, a conditional independence test can be applied to discover causality after intervention. Each noisy imputation is an independence experiment $\xi$, and the utility function becomes $U(X_{-I}|do(X_I=x_I))$. By leveraging the Monte Carlo estimator \cite{Kugelgen2019Optimal} and incorporating Assumption \ref{assump:aggr_inv}, we can estimate Eq.(\ref{eq:optim})
by:
\begin{equation}\label{eq:rewrite}
  x_I,A_c = arg\max_{x_I,A}\sum_{A\in\mathcal{A}}[P(A,X_{-I})\log P(A|X_{-I},do(X_I=x_I))].
\end{equation}
We can express Eq.(\ref{eq:rewrite}) in terms of entropy, where we have the expression as Eq.(\ref{eq:estimated}).

To solve the optimization problem above, we introduce two penalties related to the terms in Eq.(\ref{eq:estimated}):

\textit{In the first step}, for the entropy in the first term, we aim to maximize it to enhance the information gain and improve the description of causality in the inferred causal graph. This entropy can be calculated by considering the entropy of the neighbors conditioned on the intervened nodes, as expressed in Eq.(\ref{eq:int_dist}). To optimize this term, we propose a causality estimator as follows.

\begin{definition}
Given a intervention $I\in\mathcal{I}$ for the graph $G(X,A)$, the cause-effect significance of a center node $X_i$ to its neighbors $X_i\rightarrow X_j$ in GNNs can be measured by:
\begin{equation}\label{eq:ce}
  \delta\mathbf{H}(i,j)=\left\vert\mathbf{H}\left(X_j|X_i\right)-\mathbf{H}\left(X_i|X_j\right)\right\vert, i\in I, j\in Pa_i^{A},
\end{equation}
where $\mathbf{H}(\cdot|\cdot)$ is the conditional entropy.
\end{definition}

The node pair $(i,j)$ a high value of $\delta(i,j)$ implies prominent cause-effect relationships in the graph, and also a high probability of heterophily. Specifically, $X_j$ is a \textit{cause} of $X_i$ if it is positive inside the absolute-value sign. These edges are transformed into directed edges, while the opposite direction, represented by $X_j\rightarrow X_i$ in the adjacency matrix, is set to zero. It is a heuristic with a greedy strategy to learn causal structure iteratively, during which the value of the first term in Eq.(\ref{eq:estimated}), which is equivalent to maximizing ATE defined in Eq.(\ref{eq:ate}).

\textit{In the second step}, we seek to minimize the entropy of the second term of Eq.(\ref{eq:estimated}). In link prediction, a low entropy of the adjacency matrix indicates stability in the node embedding space. During the training of GNNs, we incorporate a distance penalty in the loss function to ensure consistency. This penalty ensures that the augmentation introduced by the causal structure does not disrupt the node representation, maintaining stability in the learned representations.

When the two terms in Eq.(\ref{eq:estimated}) are optimized simultaneously, the model can approach the optimal indicated by Eq.(\ref{eq:optim}).

\subsection{Insight into heterophily}
\begin{figure}[t]
  \centering
  \includegraphics[width=1\linewidth]{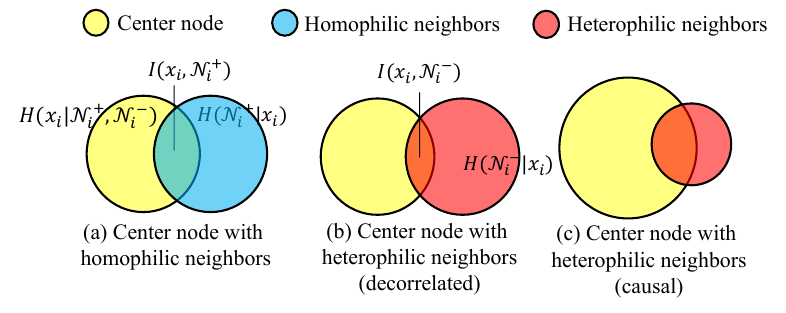}
  \caption{Venn diagram of center node and its neighbors.}
  \Description{Venn diagram of center node and its neighbors}
  \label{fig:venn}
\end{figure}

We construct a criterion for heterophilic links based on dependency estimated by their conditional entropy. Additionally, we provide insight into how conditional entropy quantifies heterophily. As a first step, we assume that the connections between node pairs are label-dependent. The expectation of a node $x_i$ connecting to a heterophilic neighbor is given by the Rayleigh quotient of the label, where the expected value is denoted as $\mathbb{E}[P_i^+]=1-R_h$ \cite{lei2022evennet}.

If we adopt a GNN as a node embedding model $f$, it learns the conditional probability distribution that relates each node to the context $x_{\mathcal{N}_i}|x_i$, where the neighbors can be divided into homophilic ones and heterophilic ones according to their labels $\mathcal{N}_i = {\mathcal{N}_i^+,\mathcal{N}_i^-}$. Then the conditional entropy is described as:
\begin{equation}\label{eq:entropy_relation}
  H(\mathcal{N}_i^-|x_i)+H(x_i|\mathcal{N}_i^-) = H(x_i,\mathcal{N}_i^+,\mathcal{N}_i^-)-H(\mathcal{N}_i^+|x_i)-I(x_i,\mathcal{N}_i^-)
\end{equation}

In heterophilic graphs, the lower bound of mutual information between heterophilic node pairs is negatively correlated with $P_i^+$ \cite{mostafa2021local}. This suggests that a heterophilic link refers to less mutual information, specifically a smaller $I(x_i,\mathcal{N}_i^-)$. Furthermore, the joint entropy $H(x_i,\mathcal{N}_i^+,\mathcal{N}_i^-)$ and the conditional entropy on the homophilic pair $H(\mathcal{N}_i^+|x_i)$ remain the same due to the static embedding model. Then, the conditional entropy on the left-hand side of the equation becomes larger.

The conditional entropy measures how well it can predict the heterophilic neighbors. When comparing the conditional entropy of a heterophilic pair to that of a homophilic pair, two situations arise. If the node pairs are less dependent on each other, both of the conditional entropy terms on the right-hand side of Eq.\ref{eq:entropy_relation} increase, resulting in a small difference between them, as Fig.\ref{fig:venn}(b). However, if there exists a symmetric dependency relationship, only one of the terms increases, leading to a large difference between $H(\mathcal{N}_i^-|x_i)$ and $H(x_i|\mathcal{N}_i^-)$, as Fig.\ref{fig:venn}(c). In the case of homophilic node pairs, the difference in conditional entropy between them remains small due to a similar representation distribution under label-dependent connectivity. Consequently, a large difference in conditional entropy suggests heterophily.

\section{Proposed model}
\begin{figure*}[t]
  \centering
  \includegraphics[width=0.7\linewidth]{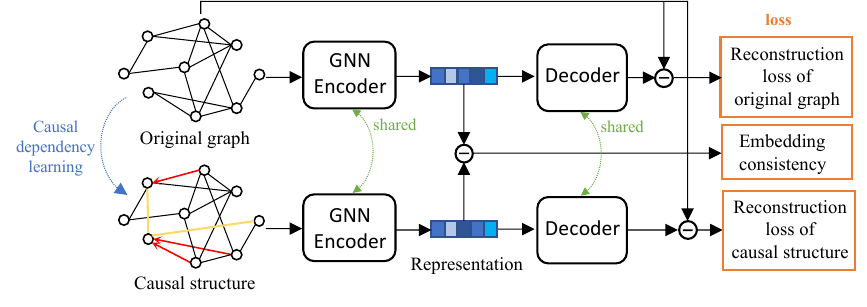}
  \caption{Main scheme of CausalMP. In each iteration, we modify the detected dependencies into directed edge (red) and add edges (yellow) through mutual information. Both graphs are encoded and decoded by GNN with shared parameter that optimized by the weighed summation of three losses.}
  \Description{Main scheme of CausalMP}
  \label{fig:main_scheme}
\end{figure*}

Based on the analysis in the previous section, we present the CausalMP to tackle the heterophily in graphs on the link prediction task. The main architecture is shown in Fig.\ref{fig:main_scheme}.

\subsection{Main modules}
\textbf{1. Intervention on graph data.} To learn the dependency among the nodes from the observational data, we employ a node feature intervention approach, where Gaussian noise rescales the node features $X$. $r_cN$ nodes are rescaled by Gaussian noise, where $r_c$ is ratio of intervened nodes. According to the \cite{peters2016causal}, it is sufficient to identify the causal connections under the noise intervention. The graph after the intervention is denoted by $G_\mathcal{I}(\hat{X},A)$, where $\hat{X}$ is the intervened feature matrix.

\textbf{2. Embedding learning.} To identify node dependencies and capture causal-effect relationships, we conduct an independence test on the distribution of observational data following the intervention. To mitigate sensitivity to downstream task correlations, we employ an unsupervised GNN denoted as $f: G\rightarrow B\in\mathcal{R}^{N\times D_{\text{emb}}}$, where $D_{\text{emb}}$ represents the dimension of the node embedding. Unsupervised learning enables the representations to capture the underlying causality among the nodes instead of correlation with the output.

\textbf{3. Node dependency estimation and causal structure modification.} We perform $M$ intervention experiments on the original graph. The trained embedding network $f$ maps the intervened graphs $\{G_\mathcal{I}[m]\}, m\in [M]$ to the embedding space $\{B_\mathcal{I}[m]\}$. Utilizing these $M$ discrete observations, we can quantify the dependency between the center nodes (i.e., intervened nodes) and their neighbors, which is calculated by Eq.(\ref{eq:ce}). In the Monte Carlo experiment, we discretize the embeddings into bins and use kernel density estimation (KDE) to estimate the joint probability density function (PDF). By leveraging the KDE estimates, we compute the conditional entropy between the node pairs.

For each edge, we obtain a corresponding dependency score $\delta\mathbf{H}(i,j)$, where a larger value implies a more prominent causal-effect relationship. For a detected dependency $X_i\rightarrow X_j$, we convert the original undirected edge $(i,j)$ to directed by setting $A(j,i)$ to 0. The threshold for pruning is  $\mu_{\mathbf{H}}+\lambda_1\sigma_{\mathbf{H}}$. Here, $\lambda_1$ is a coefficient, $\mu_{\mathbf{H}}=\mathbb{E}_{i\in\mathcal{I},j\in \mathcal{N}(i)}[\delta\mathbf{H}(i,j)]$,$\sigma_{\mathbf{H}}=\text{Var}_{i\in\mathcal{I},j\in \mathcal{N}(i)}[\delta\mathbf{H}(i,j)]$ are mean and variance of the dependency scores.

Similarly, we can  examine the presence of triangular relationships within the graph. We calculate the mutual information (MI) between the node pairs $MI_\mathcal{I}(i,j),k\in_\mathcal{I},i,j\in Pa_A(k)$. A large MI suggests a potential edge. The edges are added when MI exceeds the threshold $\mu_{\text{MI}}+\lambda_2\sigma_{\text{MI}}$, where $\mu_{\text{MI}},\sigma_{\text{MI}}$ represent the mean and variance all the MI scores, $\lambda_2$ is a coefficient.

\textbf{4. Optimization target.} After modifying the graph structure $A$, we obtain a causal structure $A_c$ . As shown in Fig.\ref{fig:main_scheme}, we employ two encoder-decoder frameworks to address the link prediction task for $A$ and $A_c$ separately. We denote the GNN as $g(G)=\text{DE}(\text{EN}(G))$, where $\text{EN}(\cdot),\text{DE}(\cdot)$ are the encoder and decoder respectively. The reconstruction loss $\mathcal{L}_{\text{recon}}$ is evaluated by the binary cross entropy between output and input logits. The consistency penalty is quantified by the mean squared error (MSE) as $\mathcal{L}_{\text{cons}}=\text{MSE}(\text{EN}(A_c),\text{EN}(A))$. The optimization target is weighted summation of the reconstruction loss of two graphs and consistency penalty $ \mathcal{L}=\mathcal{L}_{\text{recon}}(A)+\alpha\mathcal{L}_{\text{recon}}(A_c)+\beta\mathcal{L}_{\text{cons}}$, where $\alpha,\beta$ are the coefficients.

\subsection{CausalMP} 

In graph data, intervention $\mathcal{I}$ is applied to a subset of nodes. Additionally, the pruning strategy during node dependency estimation is a greedy approach. To address these limitations, we introduce iterations in the intervention experiments. It ensures that the selected center nodes and their neighbors encompass a significant portion of the nodes in the graph, then more node dependencies can be detected. A detailed algorithm of the proposed CausalMP is shown in Algorithm \ref{alg:clp}.

\begin{algorithm}[b]
  \caption{CausalMP: Causal message-passing for heterophilic graph learning.}
  \label{alg:clp}
  \KwIn{ Graph $G(X,A)$, intervention strategy $\mathcal{I}$, number intervention experiment $M$, iteration of intervention experiment $T$.}
  \KwOut{Embedding model $f$, link prediction model $g$, causal structure of input graph $A_c$.}
  Randomly initialize all the GNN models\;
  Initialize the causal structure $A_c^{(0)}=A$\;
  Train the unsupervised embedding model $f$ on $G$\;
  Pre-train $g$ on $G(X,A), G(X, A_c^{(0)})$\;
  \For{Iteration $t = 0$ to $t=T-1$}{
  Random sample $N_I$ nodes\;
  Impute noise intervention $\mathcal{I}$ on $A_c^{(t)}$ by $M$ times\;
  Map the intervened graphs to the embedding space by $f: A_c^{(t)}[M]\rightarrow B^{(t)}[M]$\;
  Estimate the PDF of node embeddings $P(B^t)$ by KDE\;
  Determine directed edges by $\delta$ threshold in Eq.(\ref{eq:ce})\;
  Determine added edges by MI threshold\;
  Update causal structure $A_c^{(t)}$ to $A_c^{(t+1)}$ \;
  Optimize $g$ on $G(X,A),G(X,A_c^{(t+1)})$\;
  }
\end{algorithm}

\subsection{Complexity analysis}
Consider a graph with $N=|V|$ nodes and $|E|$ edges. During the intervention, the number of center nodes $K$ is proportional to the node number, which is $O(|V|)$. Similarly, the number of edges involved in the dependency test would be $O(|E|)$. Then the time complexity of KDE would be $O(|E|\cdot M^2)$. For the training of the embedding network and the encoder of the link prediction model, assuming no additional tricks or computations, the time complexity would be approximately $O(N_{\text{layer}}d^2_{\text{hidden}}|V|+N_{\text{layer}}d_{\text{hidden}}|E|)$, where $N_{\text{layer}}$ is the layer number and $d_{\text{hidden}}$ is dimension of hidden layer. If we employ inner-product for the decoder, it would take $O(|V|^2)$, while $O(N_{\text{layer}}d_{\text{hidden}}|E|)$ for MLP decoder. Thus, the overall time complexity of CausalMP would not exceed $O(|V|^2+|E|)=O(N^2)$. Compared to multi-view augmentation in graph learning, the overhead of CausalMP mainly results from the KDE estimation process.

\section{Experiment}
\subsection{Experiment setup}
To evaluate the improvement over GNNs on message-pathing, we conduct a comparative analysis on link prediction task. We adopt 85\% /5\%/10\% split for training, validation, and testing. The negative link sampling capacity is the same as the positive ones. CausalMP is compared to popular baselines and SOTA models, including GAT \cite{velickovic2017graph}, VGAE \cite{kipf2016}, Graph-InfoClust (GIC)  \cite{mavromatis2021graph} and Linkless Link Prediction (LLP) \cite{guo2023linkless}.  We also compare CausalMP with two additional models specifically designed for heterophilic graphs, namely LINKX \cite{lim2021large} and DisenLink \cite{zhou2022link}, as well as another causality-based model, Counterfactual Link Prediction (CFLP)\cite{zhao2022learning}.

For the obtained explicit causal structure, we compared it with original graph in node classification. To amplify the contribution of structural information, we adopt the limited label setting as \cite{sun2023all}. Specifically, we used C-way 5-shot for small graphs and 100-shot for large graphs. We conduct the comparison on baselines models, i.e. GCN, GAT, and SOTA models for heterophilic graphs, i.e. LINKX, and GREET \cite{liu2023beyond}. As graph prompt learning also shows superiority in few-shot heterophilic node classification setting, we also apply the causal structure to GPPT \cite{sun2022gppt} and Gprompt \cite{sun2023graph}.

We conduct experiments on 9 commonly used graph datasets, encompassing 4 homophilic graphs and 5 heterophilic graphs. For the implementation of CausalMP, we utilize CCA-SSG \cite{zhang2021canonical} with default setting as the unsupervised embedding learning network. The encoder of the CausalMP is the same size as the embedding network, and we employ an MLP decoder both tasks. The details of datasets and experiment settings are shown in Appendix \ref{app:exp}. The link prediction performance is evaluated by AUC(\%) and node classification by accuracy (\%). We report the the average and variance of results for 5 experiments.

\subsection{Result}
The results of link prediction are summarized in Table \ref{tb:main_exp}, where the \textbf{bolded} and \underline{underlined} entries represent the best and second-best performance, respectively. The term "OoM" refers to out-of-memory issues of the device. Our observations are as follows: 
(a) The prevalent models and SOTA link prediction models (GIC, LLP) generally exhibit satisfactory and stable performance on homophilic graphs. However, they are unable to perform well and stably on heterophilic graphs.
(b) Models specifically designed for heterophily (LINKX, DisenLink) sacrifice their superiority on homophilic graphs, indicating a trade-off in performance depending on the graph type.
(c) CFLP demonstrates the applicability of causal analysis in heterophilic scenarios. However, it suffers from computational complexity.
(d) The proposed CausalMP exhibits stable performance and outperforms the benchmarks on both homophilic and heterophilic graphs. Notably, it remains applicable even on large graphs (CS, Physics), showcasing its scalability.

\begin{figure}[tb]
\centering
\includegraphics[width=\linewidth]{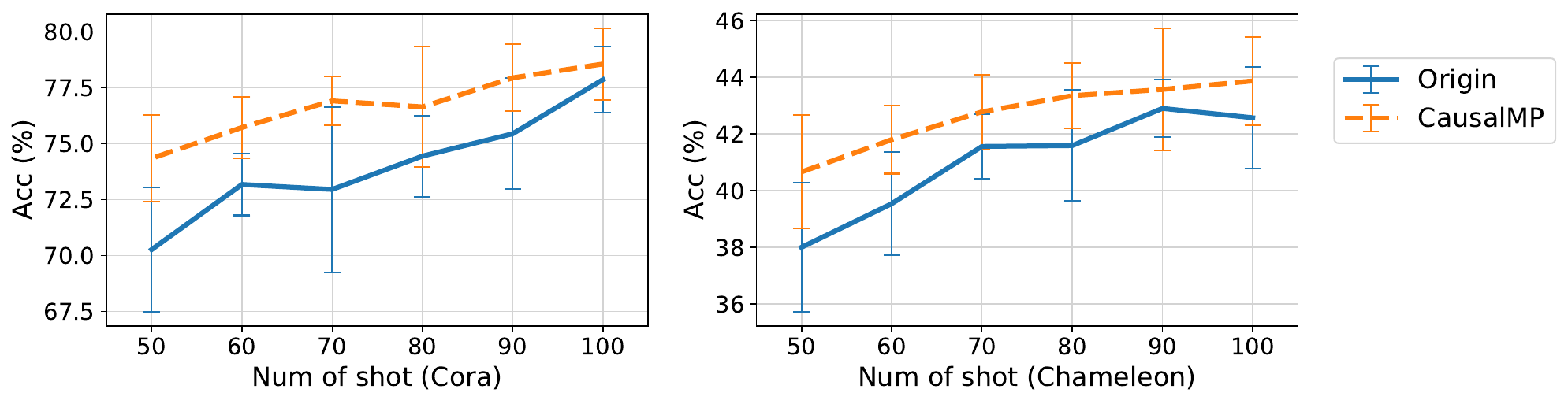}
\caption{Influence of shot number in node classification.}
\Description{Influence of shot number in node classification}
\label{fig:shot}
\end{figure}

The results of node classification with causal structure are presented in Table \ref{tb:node}. It demonstrates that the models perform better when trained with the causal structure, both for homophilic and heterophilic graphs. It indicates that the improved message-passing facilitated by the causal structure improves node representation learning by GNNs. To provide insight into the node classification experiment, we vary the number of shots and  the performance of the GCN is shown in Fig.\ref{fig:shot}, which demonstrate that the learned causal structure contributes more when there is less available node information.

\begin{table*}[t]
\centering
\caption{Performance comparison on link prediction (AUC/\%)}
\resizebox{0.9\linewidth}{!}{
\begin{tabular}{c|cccc|ccccc}
\toprule
  & \multicolumn{4}{c|}{\textbf{Homophilic}} & \multicolumn{5}{c}{\textbf{Heterophilic}}  \\
\textbf{Model} & \textbf{Cora}  & \textbf{CiteSeer} & \textbf{CS} & \textbf{Physics}  & \textbf{Actor} & \textbf{Cornell}  & \textbf{Texas} & \textbf{Chameleon}  & \textbf{Squirrel} \\ \midrule
\textbf{GAT}  & 95.14±0.57  & 96.22±0.47  & \underline{98.53±0.09}  & 97.69±0.09  & 67.80±1.12  & 61.13±3.23  & 65.73±5.06  & 97.82±1.13  & 97.03±0.16  \\
\textbf{VGAE}  & 94.78±0.69  & 95.50±0.32  & 96.65±0.14  & 94.87±0.11  & 70.82±0.81  & 58.18±9.47  & 66.75±10.09 & 98.18±0.22  & 96.59±0.24  \\
\textbf{GIC}  &  \underline{96.17±0.45}  & \underline{97.12±0.24}  & OoM  & OoM  & 70.29±0.29  & 58.01±3.41  & 66.19±7.32  & 95.30±0.29  & 95.00±0.23  \\
\textbf{LLP}  & 95.29±0.19  & 95.14±0.36  & 97.48±0.26  & \textbf{98.79±0.05}  & 80.37±1.07  & 68.20±7.96  & 71.88±3.95  & 97.52±0.37  & 95.13±0.44  \\
\textbf{LINKX} & 88.38±0.36  & 88.74±0.86  & 93.28±0.16  & 93.58±0.38  & 72.13±1.04  & 59.43±4.17  & 71.92±3.82  & 97.77±0.31  & 97.76±0.13  \\
\textbf{DisenLink}  & 89.30±0.59  & 93.96±0.88  & OoM  & OoM  & 59.19±0.48  & 60.71±5.10  & \underline{77.88±4.03}  & \underline{98.49±0.08}  & 95.88±0.10  \\
\textbf{CFLP}  & 93.44±0.82  & 93.82±0.56  & OoM  & OoM  & \underline{80.41±0.32}  & \underline{73.14±5.42}  & 66.02±3.84  & 98.29±0.16  & \textbf{98.39±0.04} \\
\textbf{CausalMP} & \textbf{96.84±0.43} & \textbf{97.20±0.43} & \textbf{98.81±0.03} & \underline{98.18±0.02} & \textbf{86.81±0.55} & \textbf{73.59±5.38} & \textbf{79.26±5.38} & \textbf{99.03±0.13} & \underline{98.11±0.15}  \\ \bottomrule
\end{tabular}\label{tb:main_exp}
}
\end{table*}

\begin{table*}[t]
\caption{The contribution of causal structure in node classification}\label{tb:node}
\centering
\resizebox{0.65\linewidth}{!}{
\begin{tabular}{cccccccc}
\toprule
  & \multicolumn{1}{c|}{} & \multicolumn{4}{c|}{\textbf{Heterophilic}}   & \multicolumn{2}{c}{\textbf{Homophilic}} \\
  & \multicolumn{1}{c|}{} & \textbf{Cornell}  & \textbf{Texas}  & \textbf{Chameron} & \multicolumn{1}{c|}{\textbf{Squirrel}} & \textbf{Cora} & \textbf{CiteSeer} \\ \hline
\multirow{2}{*}{\textbf{GCN}} & \multicolumn{1}{c|}{Original} & 46.07±2.26  & 55.49±2.28  & 41.59±1.96  & \multicolumn{1}{c|}{32.14±1.91}  & 77.87±1.48  & 71.14±1.19  \\
  & \multicolumn{1}{c|}{CausalMP} & \textbf{48.25±1.12} & \textbf{56.96±1.51} & \textbf{43.35±1.16} & \multicolumn{1}{c|}{\textbf{34.56±1.38}} & \textbf{79.15±1.16} & \textbf{72.72±1.25} \\
\multirow{2}{*}{\textbf{GAT}} & \multicolumn{1}{c|}{Original} & 45.34±2.15  & 55.71±3.10  & 39.17±1.38  & \multicolumn{1}{c|}{33.58±0.82}  & 75.41±4.88  & 69.46±2.92  \\
  & \multicolumn{1}{c|}{CausalMP} & \textbf{47.35±2.52} & \textbf{58.11±1.14} & \textbf{42.60±1.86} & \multicolumn{1}{c|}{\textbf{35.01±1.33}} & \textbf{78.25±2.81} & \textbf{71.14±1.40} \\
\multirow{2}{*}{\textbf{GREET}} & \multicolumn{1}{c|}{Original} & 57.96±4.51  & 64.56±3.46  & 52.56±2.10  & \multicolumn{1}{c|}{36.64±1.00}  & 80.32±0.86  & 71.74±0.93  \\
  & \multicolumn{1}{c|}{CausalMP} & \textbf{62.13±3.77} & \textbf{68.23±2.90} & \textbf{55.13±1.65} & \multicolumn{1}{c|}{\textbf{38.47±1.28}} & \textbf{82.20±0.97} & \textbf{72.94±0.63} \\
\multirow{2}{*}{\textbf{LINKX}} & \multicolumn{1}{c|}{Original} & 56.40±4.19  & 62.08±4.72  & 55.89±1.02  & \multicolumn{1}{c|}{37.59±0.92}  & 81.90±1.43  & 71.87±1.39  \\
  & \multicolumn{1}{c|}{CausalMP} & \textbf{59.09±3.63} & \textbf{65.76±3.27} & \textbf{57.36±0.60} & \multicolumn{1}{c|}{\textbf{39.21±0.46}} & \textbf{83.44±1.58} & \textbf{72.92±1.20} \\
\multirow{2}{*}{\textbf{GPPT}}  & \multicolumn{1}{c|}{Original} & 54.96±2.65  & 64.56±3.88  & 54.49±1.36  & \multicolumn{1}{c|}{36.16±0.98}  & 76.79 ±0.92 & 66.56±1.71  \\
  & \multicolumn{1}{c|}{CausalMP} & \textbf{57.28±3.20} & \textbf{66.78±2.90} & \textbf{56.98±1.11} & \multicolumn{1}{c|}{\textbf{37.71±0.35}} & \textbf{78.16±0.84} & \textbf{68.62±1.15} \\
\multirow{2}{*}{\textbf{Gprompt}} & \multicolumn{1}{c|}{Original} & 55.62±1.42  & 49.78±3.45  & 55.17±1.41  & \multicolumn{1}{c|}{37.14±0.80}  & 77.35 ±0.75 & 70.35±1.22  \\
  & \multicolumn{1}{c|}{CausalMP} & \textbf{59.14±2.42} & \textbf{52.78±2.17} & \textbf{57.23±1.02} & \multicolumn{1}{c|}{\textbf{39.78±0.91}} & \textbf{78.85±0.50} & \textbf{72.69±1.78} \\ \hline
\multicolumn{2}{c}{\textbf{IMP(\%)}}   & 2.82  & 2.74  & 2.30  & 1.92   & 1.74  & 1.65  \\
\multicolumn{2}{c}{\textbf{Training ratio (\%)}} & 20  & 27  & 44  & 10   & 26  & 18 \\
\bottomrule
\end{tabular}
}
\end{table*}

\subsubsection{Coefficients of the loss function} 

In order to assess the impact of the coefficients $\alpha$ and $\beta$ in the optimization target, we conducted an ablation experiment, where we modified the loss function as:
\begin{equation}
  \mathcal{L}=(1-\alpha)\mathcal{L}_{\text{recon}}(A)+\alpha\mathcal{L}_{\text{recon}}(A_c)+\beta\mathcal{L}_{\text{cons}}  
\end{equation}
where $\alpha$ represents the weight of the reconstruction loss, and $\beta$ represents the weight of the consistency between the representation of the original graph and the causal structure. 

\subsection{Ablation Experiment}
\textbf{Center node ratio.} In CausalMP, the sampling ratio of center node  $r_c \in (0,1]$ is an important parameter in the intervention strategy. We investigate its impact on two heterophilic graphs (Actor, Cornell) with $r_c \in [0,0.5]$. We report the results of the performance and corresponding time consumption in Table \ref{tb:rc}. Our findings indicate that when the iteration number is fixed, larger values of $r_c$ lead to increased time consumption, particularly on larger graphs. Excessively large values of $r_c$ can result in a more dramatic modification of the structural information and performance degradation. Conversely, when $r_c$ is too small, there are not sufficient node dependencies uncovered to improve the message-passing. To strike a balance, we experientially select $r_c$ from $[0.02, 0.1]$.

\begin{table}[t]
\caption{The influence of the center node ratio}\label{tb:rc}
\centering
\resizebox{0.9\linewidth}{!}{\begin{tabular}{ccc|ccc}
\toprule
\multicolumn{3}{c|}{\textbf{Cornell}} & \multicolumn{3}{c}{\textbf{Actor}} \\
\textbf{Rc} & \textbf{AUC(\%)}  & \textbf{Time(s)} & \textbf{Rc} & \textbf{AUC(\%)}  & \textbf{Time(s)} \\ 
\midrule
\textbf{0.05} & 79.14±5.38& 241  & \textbf{0}  & 81.74±0.54& 173  \\
\textbf{0.1}  & 79.26+5.38& 247  & \textbf{0.02} & 87.02±0.44& 658  \\
\textbf{0.2}  & 79.81±5.81& 240  & \textbf{0.05} & \textbf{87.03±0.44} & 806  \\
\textbf{0.3}  & \textbf{80.12±5.52} & 249  & \textbf{0.1}  & 86.99±0.43& 1048 \\
\textbf{0.4}  & 80.10±6.05& 261  & \textbf{0.2}  & 86.98±0.41& 1513 \\
\textbf{0.5}  & 79.55±5.40& 265  & \textbf{0.3}  & 86.93±0.42& 1965 \\ \bottomrule
\end{tabular}}
\end{table}

\begin{figure}[t]
\centering
\includegraphics[width=\linewidth]{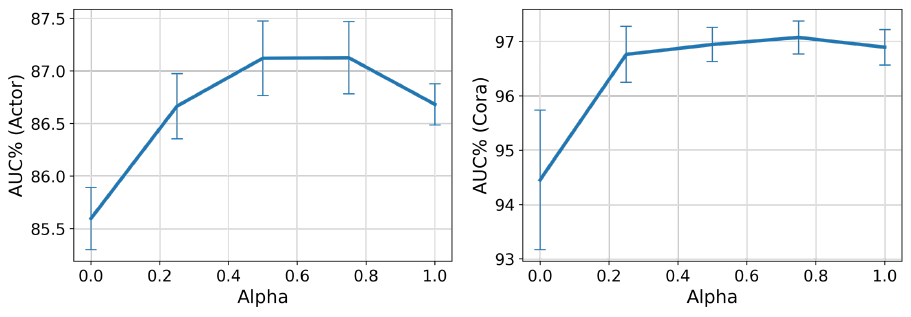}
\caption{Influence of consistency term $\alpha$.}
\Description{}
\label{fig:alpha}
\end{figure}

\begin{figure}[t]
\centering
\includegraphics[width=\linewidth]{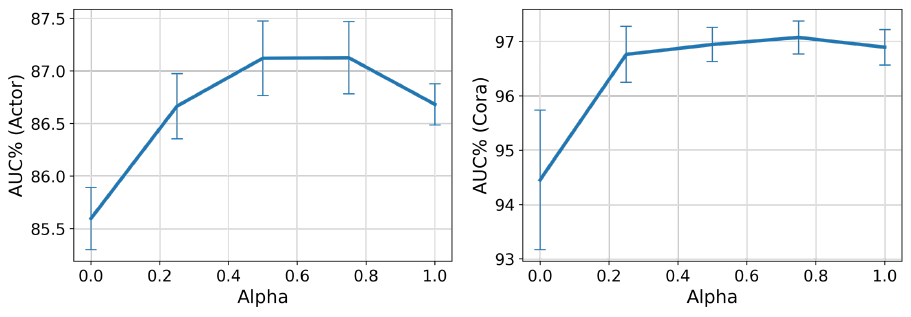}
\caption{Influence of consistency term $\beta$.}
\Description{}
\label{fig:beta}
\end{figure}

\textbf{Optimization target weight.} We first tune $\alpha$ through grid search in $[0, 0.25, 0.5, 0.75, 1]$ with result shown in Fig.\ref{fig:alpha}. We discovered that assigning a too-large weight to either the original graph or the causal graph is not beneficial for training. Optimal performance is achieved when there is a balance between the two components. Similarly, the impact of $\beta$ is evaluated in Fig.\ref{fig:beta}. We observe that as $\beta$ starts to increase from 0, the performance of CausalMP improves. This demonstrates the effectiveness of the consistency term. However, if $\beta$ keeps increasing, the performance starts to degrade. This is because an excessively large weight on the consistency term can interfere with the optimization of the reconstruction loss. While, CausalMP can achieve a stable and satisfying performance as long as extreme values of $\alpha, \beta$ are avoided. Thus, we set them as constant across different datasets.

\textbf{Edge modification strategy.} 
We conducted a comparative analysis by comparing several settings: \textit{Edge\_Aug} replaces the causal structure with an EdgeDrop augmentation. \textit{CausalMP-MI} disabled the cause-effect detection component, only retaining the adding edges strategy based on mutual information. \textit{CausalMP-$\delta$} disabled the edge addition, only retaining the proposed dependency discovery strategy. \textit{CausalMP} represents the full application of the proposed CausalMP. The link prediction results presented in Table \ref{tb:aug} reveal that both edge addition based on mutual information and node dependency discovery are effective. Combining them into full CausalMP demonstrates superior performance, improving the graph learning of GNNs.

\begin{table}[t]
\caption{Graph update strategy}\label{tb:aug}
\centering
\resizebox{\linewidth}{!}{
\begin{tabular}{ccccc}
\toprule
\textbf{Method}  & \textbf{Edge\_Aug} & \textbf{CausalMP-MI} & \textbf{CausalMP-$\delta$} & \textbf{CausalMP} \\ 
\midrule
\textbf{Cora} & 94.72±0.50 & 96.57±0.40 & 96.47±0.38 & \textbf{96.84±0.43} \\
\textbf{CS} & 97.96±0.02 & 97.95±0.02 & 98.79±0.03 & \textbf{98.81±0.03} \\
\textbf{Texas} & 72.29±5.90 & 72.68±5.68 & 72.58±5.90 & \textbf{73.59±5.83} \\
\textbf{Chameleon} & 98.76±0.10 & 98.93±0.12 & 98.91±0.09 & \textbf{99.03±0.13} \\
\textbf{Squirrel}  & 97.03±0.07 & 97.72±0.13 & 97.82±0.11 & \textbf{98.11±0.15} \\ 
\bottomrule
\end{tabular}
}
\end{table}

\textbf{Case study}
To gain a deeper understanding of how causal inference contributes to heterophilic graph learning, we conduct a case study on Texas to visualize the details. Texas is a highly heterophilic graph with 289 heterophilic edges and 36 homophilic edges. 

As first train an unsupervised node embedding network $f$ in CausalMP, we here calculate node dependency metric by Eq.\ref{eq:ce} on all the edges. 
We firstly only train the node embedding network $f$ and calculate the node dependency metric by Eq.(\ref{eq:ce}) on all the edges.
The score on the homophilic edges is $\delta_0=0.1534\pm0.1388$, while $\delta_1 = 0.2434\pm0.1894$ on the heterophilic edges. We have 99.9\%  confidence that the mean values of $\delta_0$ and $\delta_1$ are different by Z-test.

Then, in the 5 iterations of causal structure learning, the numbers of edges transferred to directed edges are [6,7,8,4,3]. [6,7,8,4,1] of them are heterophilic edges, showing that $\delta$ defined in Eq.(\ref{eq:ce}) is an effective indicator for heterophily. 

Finally, We learn a causal structure $A_c^{(T)}$ of the original graph $A$. If we re-initialize everything and pre-train a new CausalMP on $G(X,A_c^{(T)}),G(X,A_c^{(T)})$, the AUC becomes 83.98\%, while 81.82\% on the original graph $G(X,A),G(X,A)$, which shows that the efficacy of message passing is improved.

\section{Related work} 
\textbf{Heterophilic graph learning.} 
Plenty of GNNs have been proposed to tackle the heterophilic graph. The prevalent models involved in improving the information gathering, such as MixHop \cite{abu2019mixhop} and GPR-GNN\cite{chien2020adaptive}. LINKX \cite{lim2021large} learns node feature and adjacency information separately, then concatenates them for final prediction, trained through simple minibatching. GOAT \cite{kong2023goat} adaptive learns the relationship from virtually fully-connected nodes. Another promising approach is to improve the message-passing \cite{zhou2022greto,wang2022acmp,zheng2023finding}.  GReTo \cite{zhou2022greto} performs signed message passing by local context and target information. GOAL \cite{zheng2023finding} enriches the structural information by graph complementation to homophily- and heterophily-prone topology. ACMP \cite{wang2022acmp} construct message passing by interacting particle dynamics with a neural ODE solver implemented. 

\textbf{Causal inference.} 
Traditional statistical causal inference can be categorized into score-based, constraint-based methods, and hybrid methods \cite{pearl2010causal}.
It has primarily been gaining interest in the context of graph classification tasks to guide GNN training \cite{wu2022discovering, fan2022debiasing, bevilacqua2021size}, where they focus on identifying invariant substructures rather than dependency analysis. Causal inference is also applied to explanation tasks in GNN \cite{lin2021generative,lin2022orphicx}.
\cite{zhao2022learning} estimates a counterfactual adjacency matrix of the original graph to enrich the link information and improve the GNN learning. \cite{chang2023knowledge} defines the context information as the treatment and conducts augmentation on knowledge graph to improve representation learning and interpretability. \cite{Cotta2023Causal} propose a different definition of causality, namely causal lifting, and conduct the graph learning on the knowledge graph.


\section{Conclusion}
In this study, we propose CausalMP, a novel scheme with causal inference embedded, devised to deal with heterophilic graph learning by GNNs.
We conduct a theoretical analysis of intervention-based causality in GNN. We formulate an optimization problem by estimating cause-effect relationships through conditional entropy and we propose an indicator to locate heterophily.
CausalMP iteratively transfers the detected dependencies into directed edges and add edges based on mutual information, optimizing GNNs under constrastive scheme. Through extensive experiments on both homophilic and heterophilic graphs, we demonstrate that CausalMP achieves superior link prediction performance than other baselines. And the learned causal structure contributes to node classifications especially in limited label situation.

In the future, we plan to explore the inclusion of 2-hop conditions in causal analysis, striking a balance between computational complexity and accuracy. Furthermore, we intend to leverage the learned causal structure for explanation tasks.

\begin{acks}
This work is funded by National Natural Science Foundation of China Grant No. 72371271, the Guangzhou Industrial Information and Intelligent Key Laboratory Project (No. 2024A03J0628), the Nansha Key Area Science and Technology Project (No. 2023ZD003), and Project No. 2021JC02X191.
\end{acks}

\bibliographystyle{ACM-Reference-Format}
\balance
\bibliography{sample-base}

\appendix
\section{Details on Experiment }\label{app:exp}
We conduct experiments on 9 commonly used graph datasets, encompassing 4 homophilic graphs, i.e. Cora, CiteSeer, CS, Physics, and 5 heterophilic graphs, i.e. Actor, Cornell, Texas, Chameleon, and Squirrel. The dataset statistics are shown in Table \ref{tb:data}. 

\begin{table}[pht]
\centering
\caption{Dataset statistics}
\resizebox{\linewidth}{!}{
\begin{tabular}{cccccc}
\toprule
  & \textbf{\# Node} & \textbf{\# Edge} & \textbf{\# Feature} & \textbf{\# Class} & \textbf{Homo ratio} \\ \midrule
\textbf{Cora} & 2708 & 5278 & 1433 & 7  & 0.81 \\
\textbf{CiteSeer}  & 3327 & 4552 & 3703 & 6  & 0.74 \\
\textbf{CS} & 18333  & 163788 & 6805 & 15 & 0.81 \\
\textbf{Physics} & 34493  & 495924 & 8415 & {\color[HTML]{404040} 5} & 0.93 \\
\textbf{Actor}& 7600 & 33391  & 932  & 5  & 0.22 \\
\textbf{Cornell} & 183  & 298  & 1703 & 5  & 0.13 \\
\textbf{Texas}& 183  & 325  & 1703 & 5  & 0.11 \\
\textbf{Chameleon} & 2277 & 36101  & 2325 & 5  & 0.23 \\
\textbf{Squirrel}  & 5201 & 216933 & 2089 & 5  & 0.22 \\ \bottomrule
\end{tabular}\label{tb:data}
}
\end{table}

For the baseline models, GIC learns node representation through unsupervised learning by maximizing the mutual information at both the graph level and cluster level. LLP is a relational knowledge distillation framework that matches each anchor node with other context nodes in the graph. LINKX separately embeds the adjacency matrix and node feature with multilayer perceptrons and transformation. CFLP conducts causal analysis at the counterfactual level that estimates the counterfactual adjacency matrix by searching other similar node pairs in the graph under opposite contexts (community), which is a time-consuming process. For the prompt learning based models in node classification, we adopt the same implementation as \cite{sun2023all}. We adopt edge prediction in pretraining GNN with 128 hidden dimensions on small graphs (Cornell, Texas), and SimGRACE \cite{xia2022simgrace} for GNN with 512 hidden dimensions the others.

In the experiment on node dependency, the ratio of the center node in each iteration grid search within $r_c\in[0.02, 0.05, 0.1]$. We conduct an intervention for $T = 5$ iterations, repeating $M=8$ times in each iteration to estimate PDF. The parameter of the loss function is set by $\alpha=0.5, \beta=0.05$. The training epoch of the embedding network and CausalMP in each iteration are set at 1000, 2000, and learning rate 1e-4, 1e-5 respectively with a decay of 1e-4. 

The experiment is conducted on NVIDIA GeForce RTX 3090 24G GPU and Intel(R) Xeon(R) Gold 5218R CPU @ 2.10GHz.

\end{document}